\def\year{2020}\relax
\documentclass[letterpaper]{article} 
\usepackage{aaai20}  
\usepackage{times}  
\usepackage{helvet} 
\usepackage{courier}  
\usepackage[hyphens]{url}  
\usepackage{graphicx} 
\usepackage{graphicx}  
\usepackage{color}
\usepackage{subfigure}
\usepackage{amsmath}
\usepackage{amssymb}
\usepackage{verbatim}
\usepackage{bm}
\usepackage{amsfonts}
\usepackage{multirow}
\usepackage{bbm}
\usepackage{algorithm}
\usepackage{algorithmic}

\newtheorem{Definition}{Definition}
\newtheorem{theorem}{Theorem}

\DeclareMathOperator*{\argmax}{\arg\,\max}
\DeclareMathOperator*{\argmin}{\arg\,\min}
\DeclareMathOperator*{\Exp}{\mathbb{E}}

\usepackage{comment}
\title{Beyond Dropout: Feature Map Distortion to Regularize Deep Neural Networks}

\author{
	Yehui Tang,\textsuperscript{\rm 1}\thanks{This work was done while visiting Huawei Noah’s Ark Lab.}
	Yunhe Wang,\textsuperscript{\rm 2}
	Yixing Xu,\textsuperscript{\rm 2}
	Boxin Shi,\textsuperscript{\rm 4,5}
	Chao Xu,\textsuperscript{\rm 1}
	Chunjing Xu,\textsuperscript{\rm 2}
	Chang Xu\textsuperscript{\rm 3} \\
	\textsuperscript{\rm 1}Key Lab of Machine Perception (MOE), CMIC, School of EECS, Peking University, China\\
	\textsuperscript{\rm 2}Huawei Noah’s Ark Lab,
	\textsuperscript{\rm 3}School of Computer Science, Faculty of Engineering, The University of Sydney, Australia\\
	\textsuperscript{\rm 4}National Engineering Laboratory for Video Technology, Peking University
	\textsuperscript{\rm 5}Peng Cheng Laboratory\\
	\{yhtang, shiboxin\}@pku.edu.cn, \{yunhe.wang, xuyixing, xuchunjing\}@huawei.com\\ chaoxu@cis.pku.edu.cn, c.xu@sydney.edu.au
}
 
\begin{document}

\newcommand{\fracpartial}[2]{\frac{\partial #1}{\partial  #2}}
\newcommand{\norm}[1]{\left\lVert#1\right\rVert}
\newcommand{\innerproduct}[2]{\left\langle#1, #2\right\rangle}
\newcommand{\fan}[1]{\Vert #1 \Vert}
\newcommand{\qileft}{[\kern-0.15em[}
\newcommand{\qiLeft}{\left[\kern-0.4em\left[}
\newcommand{\qiright}{]\kern-0.15em]}
\newcommand{\qiRight}{\right]\kern-0.4em\right]}
\newcommand{\sign}{{\mbox{sign}}}
\newcommand{\diag}{{\mbox{diag}}}
\newcommand{\armin}{{\mbox{argmin}}}
\newcommand{\rank}{{\mbox{rank}}}
\newcommand{\<}{\left\langle}
\renewcommand{\>}{\right\rangle}
\newcommand{\lbar}{\left\|}
\newcommand{\rbar}{\right\|}
\newcommand{\eg}{{\emph{e.g. }}}
\newcommand{\ie}{{\emph{i.e.} }}
\newcommand{\wrt}{{\emph{w.r.t. }}}
\newcommand{\etc}{\emph{etc}}
\renewcommand{\algorithmicrequire}{\textbf{Input:}} 
\renewcommand{\algorithmicensure}{\textbf{Output:}} 

\renewcommand{\a}{{\bm{a}}}
\renewcommand{\b}{{\bm{b}}}
\renewcommand{\c}{{\bm{c}}}
\renewcommand{\d}{{\bm{d}}}
\newcommand{\e}{{\bm{e}}}
\newcommand{\f}{{\bm{f}}}
\newcommand{\g}{{\bm{g}}}
\newcommand{\m}{{\bm{m}}}
\renewcommand{\o}{{\bm{o}}}
\newcommand{\p}{{\bm{p}}}
\newcommand{\s}{{\bm{s}}}
\renewcommand{\u}{{\bm{u}}}
\renewcommand{\v}{{\bm{v}}}
\newcommand{\w}{{\bm{w}}}
\newcommand{\x}{{\bm{x}}}
\newcommand{\y}{{\bm{y}}}
\newcommand{\z}{{\bm{z}}}
\newcommand{\balpha}{{\bm{\alpha}}}
\newcommand{\bmu}{{\bm{\mu}}}
\newcommand{\bsigma}{{\bm{\sigma}}}
\newcommand{\blambda}{{\bm{\lambda}}}
\newcommand{\bgamma}{{\bm{\gamma}}}

\newcommand{\ba}{{\bm{A}}}
\newcommand{\bb}{{\bm{B}}}
\newcommand{\bc}{{\bm{C}}}
\newcommand{\bd}{{\bm{D}}}
\newcommand{\be}{{\bm{E}}}
\newcommand{\bg}{{\bm{G}}}
\newcommand{\bi}{{\bm{I}}}
\newcommand{\bj}{{\bm{J}}}
\newcommand{\bl}{{\bm{L}}}
\newcommand{\bp}{{\bm{P}}}
\newcommand{\bq}{{\bm{Q}}}
\newcommand{\bs}{{\bm{S}}}
\newcommand{\bu}{{\bm{U}}}
\newcommand{\bv}{{\bm{V}}}
\newcommand{\bw}{{\bm{W}}}
\newcommand{\bx}{{\bm{X}}}
\newcommand{\by}{{\bm{Y}}}
\newcommand{\bz}{{\bm{Z}}}
\newcommand{\bTheta}{{\bm{\Theta}}}
\newcommand{\bSigma}{{\bm{\Sigma}}}

\newcommand{\A}{{\mathcal{A}}}
\newcommand{\B}{\mathcal{B}}
\newcommand{\C}{\mathcal{C}}
\newcommand{\D}{\mathcal{D}}
\newcommand{\E}{\mathcal{E}}
\newcommand{\F}{\mathcal{F}}
\renewcommand{\H}{\mathcal{H}}
\newcommand{\K}{\mathcal{K}}
\renewcommand{\L}{\mathcal{L}}
\newcommand{\W}{\mathcal{W}}
\newcommand{\X}{\mathcal{X}}
\newcommand{\Y}{\mathcal{Y}}
\newcommand{\Q}{\mathcal{Q}}
\newcommand{\R}{\mathbb{R}}

\newcommand{\T}{\mathcal{T}}
\newcommand{\1}{\mathbbm{1}}

\maketitle

\begin{abstract}
	Deep neural networks often consist of a great number of trainable parameters for extracting powerful features from given datasets.  On one hand, massive trainable parameters significantly enhance the performance of these deep networks. On the other hand, they bring the problem of over-fitting. To this end, dropout based methods disable some elements in the output feature maps during the training phase for reducing the co-adaptation of neurons. Although the generalization ability of the resulting models can be enhanced by these approaches, the conventional binary dropout is not the optimal solution. Therefore, we investigate the empirical Rademacher complexity related to intermediate layers of deep neural networks and propose a feature distortion method (Disout) for addressing the aforementioned problem. In the training period, randomly selected elements in the feature maps will be replaced with specific values by exploiting the generalization error bound. The superiority of the proposed feature map distortion for producing deep neural network with higher testing performance is analyzed and demonstrated on several benchmark image datasets.
\end{abstract}

\section{Introduction}

The superiority of deep neural networks, especially convolutional neural networks (CNNs) has been well demonstrated in a large variety of computer vision tasks including image recognition~\cite{krizhevsky2012imagenet,he2016deep,wang2018learning}, object detection~\cite{ren2015faster,redmon2016you}, video analysis~\cite{feichtenhofer2016convolutional}, Natural Language Processing~\cite{wang190409408} \etc. Actually, the huge success of deep CNNs should be attributed to the larger number of trainable parameters and available annotation data, \eg the ImageNet~\cite{deng2009imagenet} dataset with over 1 million images from 1000 different categories. 

Since deep networks are often over parameterized for achieving higher performance on the training set, an important problem is to avoid over-fitting, \ie the excellent performance achieved on the train set is expected to be repeated on the test set~\cite{hinton2012improving,wang2018packing}. In other words, the empirical risk should be closed to the expected risk. To this end,~\cite{hinton2012improving} first proposed the conventional binary dropout approach, which  reduces the co-adaptation of neurons by stochastically dropping part of them in the training phase. This operation  can be either regarded as a model ensemble technique or a data augmentation method, which significantly enhances the performance of the resulting network on the test set.

To improve the performance of dropout implemented on deep neural networks, \cite{ba2013adaptive} adaptively adjusted the dropout probability of each neuron by interleaving a binary belief network into the neural networks. Gaussian Dropout~\cite{srivastava2014dropout} multiplying the outputs of the neurons by Gaussian random noise is equal to the conventional binary dropout. It was further analyzed from the perspective of Bayesian regularization and the dropout probability can be optimized automatically~\cite{kingma2015variational}. Instead of disabling the activation, DropConnect~\cite{wan2013regularization} randomly set a subset of network weights to zero. \cite{wan2013regularization} derived a bound on the generalization performance for Dropout and DropConnect. \cite{zhai2018adaptive} connected the bound with drop probability and optimized the dropout probability together with network parameters during the training. Focusing on the convolutional neural networks, ~\cite{ghiasi2018dropblock} proposed to drop contiguous regions of a feature map to obstruct the information flow more radically. 

Existing variants of dropout have made tremendous efforts for minimizing the gap between the expected risk and the empirical risk, but they all follow the general idea of disabling parts of the output of an arbitrary layer in the neural network. The essence of the success is to randomly obscure part of semantic information extracted by the deep neural network and avoid the massive parameters to over-fit the training set. Setting a certain number of the elements in the feature map to zero is a straightforward way to disturb the information propagation across layers in the neural network, but it is by no means the only way to accomplish this goal. Most importantly, such sort of hand-crafted operations are hardly to be the optimal ones in most cases. 

In this work, we propose a novel approach for enhancing the generalization ability of deep neural networks by investigating the distortion on the feature maps (Disout). The generalization error bound of the given deep neural network is established in terms of the Rademacher complexity of its intermediate layers. Distortion is introduced onto the feature maps to decrease the associated Rademacher complexity, which is then beneficial for improving the generalization ability of the neural network. Besides minimizing the general classification loss, the proposed distortion can simultaneously minimize the expected and empirical risks by adding distortions on feature maps. An extension to convolutional layers and corresponding optimization details are also provided. Experimental results on benchmark image datasets demonstrate that deep networks trained using the proposed feature distortion method perform better than those generated using state-of-the-art methods.

\section{Preliminary}
Dropout is a prevalent regularization technology to alleviate over-fitting of models and has achieved great success. It has been demonstrated dropout can  improve the generalization ability of models both theoretically \cite{wan2013regularization} and practically \cite{srivastava2014dropout}. In this section, we briefly introduce  the generalization theory and dropout method.
\subsection{Generalization Theory}

Generalization theory focuses on the relation between the expected risk and the empirical risk. Considering an $L$-layer neural network $\f ^L \in \F$, and a labeled dataset $\D= \left \{(\x_i,\y_i) \right \}_{i=1}^{N}$ sampled from the ground-truth distribution $\Q \in \X \times \Y$, in which $\x_i \in \X$ and $\y_i \in \Y$. Denote the weight matrix as $\K^{l}\in \R^{d^{l}\times d^{l-1}}$ in which $d^l$ is the dimension of the feature map of $l$-th layer, and the corresponding output features before and after activation functions $\phi$ of the $l$-th layer as $\o^{l}\in \R^{d^{l}}$ and $\f^{l}\in \R^{d^{l}}$, respectively. Omitting bias, we have $\f^{l+1}(\x_i)=\phi(\o^{l+1}(\x_i))=\phi(\K^{l+1}\f^{l}(\x_i))$. For simplicity, we further refer $\K^{:l}$ as $\{\K^{1}, \cdots, \K^{l}\}$.

Taking the image classification task as an example, the expected risk $R(\f^L)$ over the population and the empirical risk $\hat R(\f^L)$ on the training set can be formulated as:
\begin{align}
\small
R(\f^L)&= \Exp_{(\x,\y) \sim \mathcal{Q}}\qileft \ell(\f^L(\x,\K^{:L}),\y)\qiright, \\
\hat{R}(\f^L)&= \frac{1}{N} \sum_{(\x_i,\y_i) \in \D} \ell(\f^L(\x_i,\K^{:L}),\y_i),
\end{align}
where $\ell(\cdot)$ denotes 0-1 loss. Various techniques have been developed to quantify the gap between the expected risk and the empirical risk, such as PAC learning ~\cite{hanneke2016optimal} , VC dimension ~\cite{sontag1998vc} and Rademacher complexity \cite{koltchinskii2002empirical}. Wherein, the empirical Rademacher complexity (ERC) has been widely used as it often leads to a much tighter generalization error bound. The formal definition of ERC is given as follows:
\begin{Definition} 
	\label{th-rad}
	For a given training dataset with $N$ instances $\D= \left \{(\x_i,\y_i) \right \}$ generated by the distribution $\Q$, the empirical Rademacher complexity of the function class of the network $\f^L$ is defined as: 
	\begin{equation}
	\small
	\tilde R_{D}(\f^{L})=\frac{1}{N} \Exp_{\bm \sigma} \left | \sup_{k,\K^{:L}} \sum_{i=1}^{N} \sigma_i \f^L (\x_i,\K^{:L} )[k] \right|,
	\end{equation}
	where Rademacher variables $\bm \sigma=\{ \sigma_1,\cdots,\sigma_N\}$, $\sigma_i$'s are independent uniform random variables in \{-1,+1\} and $\f^L (\x_i,\K^{:L})[k]$ is the $k$-th element in $\f^L (\x_i,\K^{:L})$.	
\end{Definition}
Using empirical Rademacher complexity and MaDiarmid's inequality, the upper bound of the expected risk  $R(\f^L)$ can be derived by  Theorem \ref{th-rad}~\cite{koltchinskii2002empirical}.
\begin{theorem} 	
	\label{th-bound}
	Given a fixed $\rho >0$, for any $\delta>0$, with probability at least $1- \delta$, for all $\f^L \in \F$
	\begin{align}
	\begin{split}
	R(\f^{L}) &\le \hat{R} (\f^{L})+\frac{2(d^L)^2}{\rho} \tilde R_{D}(\f^{L})\\
	&+ \left (1+\frac{2(d^L)^2}{\rho} \right)\sqrt{\frac{\ln{\frac{1}{\delta}}}{2N}},
	\end{split}
	\end{align}	
	where $d^L$ denotes the output dimension of the network.
\end{theorem}
According to Theorem \ref{th-bound} we can find that the gap between expected and empirical risks can be bounded with the help of the empirical Rademacher complexity $\tilde  R_{D}(\f)$ over the specific neural network and dataset. Directly calculating the ERC is vary hard~\cite{kawaguchi2017generalization}, and thus the upper bound or approximate values of the ERC are usually used in the training phase for obtaining models with better generalization~\cite{kawaguchi2017generalization,zhai2018adaptive}. \cite{kawaguchi2017generalization} obtained models with better generalization by decreasing a regularization term related to the ERC. The effectiveness of decreasing ERC  in previous works inspires us to leverage ERC to refine the conventional dropout methods.
\subsection{Dropout}
Dropout is a classical and effective regularization technology to improve the generalization capability of models. There are many variants of dropout,\eg variational dropout and \cite{kingma2015variational} DropBlock~\cite{ghiasi2018dropblock}). Most of them follows the technology of disabling part elements of the feature maps. In general, these methods can be formulated as:
\begin{equation}
\hat{\f}^l(\x_i) =  \f^l(\x_i) - \m^l_i \circ \bm \f^l(\x_i),  
\label{eq:drop} 
\end{equation} 
where $\circ$ denotes the element-wise product, $\f^l(\x_i)$ \footnote{Without ambiguity, $\f^l(\x_i,\K^{:l})$ is denoted as $\f^l(\x_i)$ for simplicity.} and $\hat{\f}^l(\x_i)$ are the original feature and distorted features, respectively. In addition, $\m^l_i \in \{0, 1\}^{d^l}$  is the  binary mask applied on feature map $f^l(\x_i)$, and each element in $\m^l_i$ is draw  from Bernoulli distribution, \ie set to 1 with the dropping probability $p$.  Admittedly, implementing dropout on the features in the training phase will force the given network paying more attentions on those non-zero regions, and partially solve the ``over-fitting''. However, disabling the original feature is a heuristic approach and may not always leads to the optimal solution for addressing the aforementioned over-fitting problem in deep neural networks. 
\section{Approach}
Instead of fixing the value of perturbation, we aim to learn the distortion of the feature map by reducing the ERC of the network. Generally, the disturbing operation employed on the output feature $\f^l(\x_i)$  of the $l$-th layer with input data $\x_i$ can be formulated as:
\begin{equation}
\hat{\f}^l(\x_i) =  \f^l(\x_i) - \m^l_i \circ \bm \varepsilon^l_i,  
\label{eq:dist} 
\end{equation}	
where $\bm \varepsilon^l_i \in \R^{d^l}$  is the distortion applied the on feature map $\f^l(\x_i)$. Compared to the dropout method (Eq. (\ref{eq:drop}))  which manually set the distortion as $\varepsilon^l_i=\bm \f^l(\x_i)$,  Eq. (\ref{eq:dist}) automatically learns the form of distortion in the guide of ERC.    
Directly using $\tilde R_{D}(\f^{L})$ which is the ERC of the network to guide the distortion $\bm \varepsilon^l_i$ is very hard. Since $\tilde R_{D}(\f^{L})$ is calculated on the final layer \wrt the output of the neural network, and it is difficult to trace the intermediate feature maps of the neural network  during  the training phase. 
Hence, we reformulate $\tilde R_{D}(\f^{L})$ by considering the output feature of an arbitrary layer, and obtain the following theorem based on \cite{wan2013regularization}.
\begin{theorem} 
	\label{th-trans}
	Let $\K^l[k,:]$ denotes the $k$-th row of the weight matrix $\K^l$ and $\|\cdot\|_p$ is the p-norm of vector.  Assume that $\|\K^l[k,:]\|_p \le B^l$, and then the ERC of output can be bounded by the ERC of intermediate feature:
	\begin{align}
	\begin{split}
	&\tilde R_{D}(\f^{L}) \le 2\tilde R_{D}(\o^{L}) \le 2B^L \tilde R_{D}({\f^{L-1}}) \le \cdots \\
	&\le 2^{L-t} \tilde R_D (\f^t)  \prod_{l=t+1}^{L}B^l \le 2^{L-t+1} \tilde R_D (\o^t)  \prod_{l=t+1}^{L}B^l,
	\end{split}
	\end{align}	
	where $\o^{l}$ and $\f^l$ are the feature maps before and after activation function respectively. 
\end{theorem} 
The above theorem shows that the ERC of the network $\tilde R_{D}(\f^{L})$ is upper bounded by the ERC of output feature $\tilde R_D (\f^t)$ or $\tilde R_D (\o^t)$ of $t$-th layer \footnote{The definition of $\tilde R_D (\f^t)$ and $\tilde R_D (\o^t)$ in $t$-th layer has the same form as Definition \ref{th-rad}, \ie $\tilde R_{D}(\f^{t})=\frac{1}{N} \Exp_{\bm \sigma} \left | \sup_{k,\K^{:t}} \sum_{i=1}^{N} \sigma_i \f^t (\x_i,\K^{:t} )[k] \right|$ and $\tilde R_{D}(\o^{t})=\frac{1}{N} \Exp_{\bm \sigma} \left | \sup_{k,\K^{:t}} \sum_{i=1}^{N} \sigma_i \o^t (\x_i,\K^{:t} )[k] \right|$}. Thus, decreasing $\tilde R_D (\f^t)$ or $\tilde R_D (\o^t)$ can heuristically decrease $\tilde R_{D}(\f^{L})$. Note that $\f^t$ is the feature map of arbitrary intermediate layer $t$ of the network, and the distortion is also applied on intermediate features. Thus, $\tilde R_D (\f^t)$ or $\tilde R_D (\o^t)$ is used to guide the distortion in the following.
\subsection{Feature Map Distortion}
In this section, we will illustrate the way of decreasing ERC by applying the distortion $\bm \varepsilon^l$ on the feature map of $l$-th layer $\f^l(\x_i)$. By doing so, all the ERCs in the subsequent layers will be affected, and $\tilde R_D (\o^t)$ satisfying $l < t \le L$ can guide the distortion $\bm \varepsilon^l$ of $l$-th layer. Recall that in theorem \ref{th-trans}, the closer a layer is to the output layer, the tighter the upper bound of the ERC of the whole network is, and may reduce  $\tilde R_{D}(\f^{L})$ more effectively. However, if $t \gg l$, the relationship between $ \tilde R_{D}(\o^{t})$ and $\bm \varepsilon^l$ becomes complex and it is difficult to guide $\bm \varepsilon^l$ with $ \tilde R_{D}(\o^{t})$.  
Thus, we use the ERC of $(l+1)$-th layer $\tilde R_{D}(\o^{l+1})$ to guide the distortion $\bm \varepsilon^l$ in $l$-th layer. Specifically, we reduce $\tilde R_{D}(\o^{l+1})$ by optimizing $\bm \varepsilon^l$. Denoting 
\begin{equation}
\g^l(\x)=\sum_{i=1}^N \sigma_i  \hat{\f^{l}}(\x_i), 
\end{equation}
for simplicity, $\g^l(\x)\in \R^{d^l}$ has the same dimension as feature map $\f^{l}(\x_i)$. And then, $\tilde R_{D}(\o^{l+1})$ is calculated as:
\begin{align}
\begin{split}
\tilde R_{D}(\o^{l+1}) 
=\frac{1}{N} \Exp_{\bm \sigma} \sup_{k,\K^{:l+1}} \left| \left< \K^{l+1}[k,:]^T ,\g^{l}(\x) \right > \right|,
\end{split}
\end{align}
where $\K^{l+1}[k,:] \in \R^{1 \times d^l} $ denotes the $k$-th row of the weight matrix $\K^{l+1}$ and $\K^{:l+1}=\{\K^1,\K^2,\cdots,\K^{l+1}\}$. 
An ideal $\bm \varepsilon^l$ will reduce the ERC of the next layer $\tilde R_{D}(\o^{l+1})$ while preserving the representation power.

During the training phase, considering a mini-batch $\bar\x=\{\x_1,\x_2,\cdots \x_{\bar N}\}$ with $\bar N$ samples, the mask and distortion of the $l$-th layer are $\m^l=\{\m^l_1,\m^l_2,\cdots, \m^l_{\bar N}\}$ and $\bm \varepsilon^l=\{\bm \varepsilon^l_1, \bm \varepsilon^l_2,\cdots, \bm \varepsilon^l_{\bar N}\}$, respectively. Taking the classification problem as an example, the weights of the network are updated via minimizing the cross-entropy loss. Based on the current updated weights $\K^l$ and Rademacher variables $\bar{\bm \sigma}=\{\sigma_1,\sigma_2, \cdots, \sigma_{\bar N}\}$, the optimized disturbance $\hat {\bm \varepsilon}^l$ is obtained by solving the optimization problem:
\begin{equation}
\label{eq-tl}
\hat{\bm \varepsilon}^{l}=\argmin_{\bm \varepsilon^{l}} \T(\bar\x,\bm \varepsilon^l),\ \  l=1,2,\cdots, L
\end{equation}
where
\begin{align}
\label{eq-tl2}
\begin{split}
&\T(\bar\x,\bm \varepsilon^l)\\
&=\frac{1}{\bar{N}} \left [ \sup_k \left| \left< \K^{l+1}[k,:]^T ,\g^{l}(\bar\x) \right>\right| + \frac{\lambda}{2} \sum_{i=1}^{\bar N}  \|\bm \varepsilon^l_i\|_2^2   \right] , 
\end{split}
\end{align} 
in which $\|\cdot\|_2$ denotes the $l_2$-norm of the vector and $\lambda$ is a hyper-parameter balancing the objective function and the intensity of distortion. Intuitively, a violent distortion will destroy the original feature and reduce the representation power.

\subsection{Optimization of the Distortion} 
Our goal is to reduce the first term in Eq. (\ref{eq-tl2}) related to ERC while constraining the intensity of distortion $\bm \varepsilon^l_i$. Note that the conventional dropout which sets $\bm \varepsilon^l_i=\f^l(\x_i)$ also achieves the similar goal in a special situation. When the drop probability $p=1$ and all the elements in mask $\m^l_i$ are set to 1, the distortion $\bm \varepsilon^l_i=\f^l(\x_i)$ makes $\g^{l}(\bar \x)=0$ and thus the first term in Eq. (\ref{eq-tl2}) is zero, showing that the dropout also has the potential to reduce ERC. However, the semantic information is also dropped away and the network will make random guess.
In the general case where $p < 1$,  the conventional dropout disables part of the feature maps, which may decrease the value of  $\T(\bar \x,\bm \varepsilon^l)$, but there is no explicit interaction with the empirical Rademacher complexity. We choose $\f^l(\x_i)$ as the initial value of $\bm \varepsilon^l_i$ and optimize Eq.~(\ref{eq-tl}) with gradient descent. 
The partial derivative of $\T(\bar \x,\bm \varepsilon^l)$ \wrt $\bm \varepsilon^{l}_i$ is calculated as:
\begin{equation} 
\label{eq-der}
\frac{\partial \T(\bar \x,\bm \varepsilon^l)}{\partial \bm \varepsilon^{l}_i}= -\frac{1}{\bar N}  \sigma_i s_{\hat k}\K^{l+l}[\hat k,:]^T \circ \m^l_i + \frac{\lambda}{\bar{N}} \bm \varepsilon^l_i,  
\end{equation}
where
\begin{align}
\label{eq-der2}
\hat k&= \argmax_{k}  \left|  \left< \K^{l+1}[ k,:]^T ,\g^{l}(\bar \x) \right>\right|,\\
\label{eq-der3}
s_{\hat k}&=\sign \left< \K^{l+1}[\hat k,:]^T ,\g^{l}(\bar \x) \right>.
\end{align}
Eq.~(\ref{eq-der2}) chooses the row of weight matrix  to obtain the maximum inner product $\left< \K^{l+1}[ k,:]^T ,\g^{l}(\bar \x) \right>$ and Eq.~(\ref{eq-der3}) calculates the sign of the inner product. The equations above show that the optimization of distortion $\bm \varepsilon^l$ is related to the feature $\f^l(\x_i)$ and the weight $\K^{l+1}$ in the following layer.
Note that precisely calculating the gradient $\frac{\partial \T(\x,\bm \varepsilon^l)}{\partial \bm \varepsilon^{l}_i}$ is time-consuming and not necessary, and it can be appropriately estimated without much influence of the performance. Rademacher variable $\sigma_i$ is randomly sampled from $\{\pm 1\}$ with equal probability (Definition \ref{th-rad}), and thus the impact of $s_{\hat k}$ can be neglected. Selecting the row index $k$ of $\K^{l+1}$ is also related to the random variable $\sigma_i$, and hence we leverage the random variables to approximate the process. Denote $\K^{l+1}_M=[\max(\K^{l+1}[:,1]),\max(\K^{l+1}[:,2]),\cdots,\max(\K^{l+1}[:,d^l])]^T$ in which the $j$-th element is the maximum value of the $j$-th column of weight matrix $\K^{l+1}$.
Then the gradient $\frac{\partial \T(\x,\bm \varepsilon^l)}{\partial \bm \varepsilon^{l}_i}$ is approximated as: 
\begin{equation}
\label{eq-pap}
\frac{\partial T^{l+1}}{\partial \bm \varepsilon^{l}_i} \approx - \frac{1}{\bar N}  \sigma_i \u \circ \K^{l+1}_M \circ \m^l_i + \frac{\lambda}{\bar{N}} \bm \varepsilon^l_i,  
\end{equation}
where $\u \in d^l$ is a random variable whose elements are sampled from standard normal distribution $\mathcal N(0,1)$ with zero mean and standard deviation. $\u \circ \K^{l+1}_M$ is to approximate the process of selecting the row of weight $\K^{l+1}$.
Denote $\gamma$ as the step length and we can update $\bm \varepsilon^{l}_i$ along the negative gradient direction:
\begin{equation}
\bm \varepsilon^{l}_i \leftarrow \bm \varepsilon^{l}_i - \gamma \frac{\partial T^{l+1}}{\partial \bm \varepsilon^{l}_i}.
\end{equation}
To train an optimal neural network, we tend to simultaneously reduce the empirical risk on the training dataset (\eg minimizing the cross entropy) and the Rademacher complexity. There is thus a balance between the ordinary loss and the reduction of Rademacher complexity. This can be realized by alternatively optimizing between the ordinary loss \wrt weights of the network and Rademacher complexity \wrt the distortion $\bm \varepsilon^l$. After obtaining the updated weights of the network, the distortion $\bm \varepsilon^l_i$ is optimized to decrease the objective $\T(\x,\bm \varepsilon^l)$. After each update of weights of the network, the $\bm \varepsilon^l_i$ can be updated for several times, which is usually adopted in practice for training efficiency \cite{goodfellow2014generative}. Using the case that applying distortion on feature maps of all the layers as an example,  the training procedure of the network  is summarized in Algorithm \ref{alg}. Following dropout\cite{srivastava2014dropout}, the feature map is rescaled by a factor of $p$ at testing stage, which is equally implemented as dividing $p$ in the training phase in practice\cite{srivastava2014dropout}.
\begin{algorithm}[tb]
	\caption{Feature map distortion for training networks.}
	\label{alg}
	\begin{algorithmic}[1]
		\REQUIRE{Training data $\D = \{(\x_i,\y_i)\}_{i=1}^N$}, The weights of the network $\K^{:L}=\{\K^1,\K^2,\cdots,\K^L\}$
		\REPEAT
		\FOR{ $l$ in $1, \cdots L$ }	
		\STATE Calculate the feature map $\f^l(\x_i)$ of the $l$-th layer;
		\STATE Generate the distortion $\bm \varepsilon^{l}_i$ and the corresponding sample mask $\m^l_i$;
		\STATE Obtain distorted feature $\hat \f^l(\x_i)$ (Eq.~(\ref{eq:dist}));
		\STATE Feed-forward the network using $\hat \f^l(\x_i)$;
		\ENDFOR
		\STATE Backward and update weights $\K^{:L}$ in the network;
		\UNTIL Convergence;
		\ENSURE{The resulting deep neural network.}
	\end{algorithmic}
\end{algorithm}
\begin{table*}
	
	\caption{Accuracies of conventional CNNs on CIFAR-10 and CIFAR-100 datasets.}
		\vskip 0.1in
	\label{tb-ccnnres}
	\centering
	\begin{tabular}{l||c|c}
		\hline
		Method & CIFAR-10 (\%) & CIFAR-100 (\%)\\ \hline\hline
		CNN &81.99&49.72\\
		CNN + Dropout \cite{srivastava2014dropout}& 82.95&54.19\\
		CNN + Vardrop \cite{kingma2015variational} &83.15&54.53\\
		CNN + Sparse Vardrop \cite{molchanov2017variational}&82.13&54.26\\
		CNN + RDdrop \cite{zhai2018adaptive}&83.11&54.65\\
		\hline
		CNN + Feature Map Distortion &\textbf{85.24 $\pm$ 0.08}&\textbf{56.23$\pm$ 0.12}\\ 
		\hline
		
	\end{tabular}
	\vskip - 0.1in	
\end{table*}
\subsection{Extension to Convolutional Layers}
Convolutional layer can be seen as a special full-connected layer with sparse connection and shared weights. Hence, the distortion $\bm \varepsilon^l$ can be learned in the same way as that in the FC layer. In the following, we focus on  distorting the feature maps to reduce the empirical Rademacher complexity in convolutional layers, considering the particularity of convolution operations. 

The convolutional kernel of $l$-th layer is denoted as $\K^{l}\in \R^{d^{l}_c \times d^{l-1}_c \times d^{l-1}_h \times d^{l-1}_w}$, and the corresponding output feature maps before and after activation function $\phi$ are denoted as $O^l (\x_i) \in \R^{d^l_c\times  d^l_{h'} \times d^l_{w'}}$ and $F^l (\x_i) \in \R^{d^l_c\times  d^l_{h'} \times d^l_{w'}}$, respectively. $d_h^l$ and $d_w^l$ are the height and  width of convolutional kernels while $d_{h'}^l$ and $d_{w'}^l$ are those of the feature map. The mask $M_i^l \in \R^{d^l_c\times  d^l_{h'} \times d^l_{w'}}$ and distortion $\bm \varepsilon^l_i\in \R^{d^l_c\times  d^l_{h'} \times d^l_{w'}}$ of the $l$-th layer  have the same dimension as feature map $F^l (\x_i)$ and is applied to $F^l (\x_i)$ to get the disturbed feature map $\hat{F}^l (\x_i)$, \ie
\begin{equation}
\label{eq-disconv} 
\hat{F}^l (\x_i) =  F^l (\x_i) - M^l_i \circ \bm \varepsilon^l_i. 
\end{equation}
Similar to the fully-connected layer, the ERC  $\tilde R_{D}(O^{l+1})$ in the $(l+1)$-th layer  is used to guide the optimization of distortion $\bm \varepsilon^l$ in layer $l$. Given a mini-batch $\bar\x=\{\x_1,\x_2,\cdots \x_{\bar N}\}$ together with  mask $M^l=\{M^l_1,M^l_2,\cdots, M^l_{\bar N}\}$ and distortion $\bm \varepsilon^l=\{\bm \varepsilon^l_1, \bm \varepsilon^l_2,\cdots, \bm \varepsilon^l_{\bar N}\}$, and two symbols $G^l(\bar\x)$ and $Q^{l+1}(\bar\x)$ are defined for notion simplicity:
\begin{align}
&G^l(\bar\x)=\sum_{i=1}^N \sigma_i \hat{F}^l (\bar {\x}_i), \\
\label{eq-q}
&Q^{l+1}(\bar\x)[k,:,:]=\sum_{c=1}^{d^{l}_c}  \K^{l+1}[k,c,:,:] *  G^l[c,:,:],
\end{align}
where $*$ denotes convolutional operation. $G^l(\bar\x)$ is related to the distorted feature and the Rademacher variable in the $l$-th layer, and Eq. (\ref{eq-q}) applies the convolutional operation on $G^l(\bar\x)$. Given the notation mentioned above, $\bm \varepsilon^l$ can be derived by minimizing the following objective function:
\begin{equation}
\hat{\bm \varepsilon}^{l}=\argmin_{\bm \varepsilon^{l}}
\T(\bar\x,\bm \varepsilon^l),
\end{equation}
where
\begin{align}
\scriptsize
\label{tarcnn}
\begin{split}
&\T(\bar\x,\bm \varepsilon^l)
=\frac{1}{\bar{N}d^{l+1}_{h}d^{l+1}_{w}} \sup_k \sum_{h'=1}^{d^{l+1}_{h'}} \sum_{w'=1}^{d^{l+1}_{w'}}
\left|Q^{l+1}(\bar\x)[k,h',w']\right| \\
& + \frac{\lambda}{2\bar{N}} \sum_{i=1}^{\bar N}\| \bm \varepsilon^l_i \|_2^2.
\end{split}
\end{align} 
$\T(\bar\x,\bm \varepsilon^l)$ comes from the simplified implementation $\tilde R_{D}(O^{l+1})$ which is the ERC in a mini-batch. As Eq. (\ref{tarcnn}) calculates average over the spatial dimension of $Q^{l+1}(\bar\x)$, elements in different spatial locations of $\bm \varepsilon_{i}^{l}$ has equal contribution to $Q^{l+1}(\bar\x)$. Thus, the partial derivative of $Q^{l+1}(\bar\x)$ \wrt $ \bm \varepsilon_{i}^{l}$ is:
\begin{align} 
\small
\label{eq-conv-dev}
\begin{split}
&\frac{\partial \T}{\partial \bm \varepsilon_{i}^{l}[c,h',w']}
= -\frac{1}{\bar N d^{l+1}_{h}d^{l+1}_{w}} \sigma_i \sum_{h=1}^{d^{l+1}_{h}}\sum_{w=1}^{d^{l+1}_{w}} \K^{l+1}[\hat k,c,h,w] S[\hat k,h,w] \\
&+ \frac{\lambda}{\bar N} \bm \varepsilon^l_i, h' \in \{1, 2 \cdots, d^{l}_{h'}\},\ \  w' \in \{1, 2 \cdots, d^{l}_{w'}\},
\end{split}
\end{align}
where
\begin{align}
\label{eq-conv-dev2}
&\hat k= \argmax_k \sum_{h=1}^{d^{l+1}_{h'}} \sum_{w=1}^{d^{l+1}_{w'}} \left|Q^{l+1}(\bar\x)[k,h',w'] \right |,\\
\label{eq-conv-dev3}
&S=\sign \left(Q^{l+1}(\bar\x) \right)
\end{align}
in which $S\in \{\pm 1\}^{d^l_{c} \times d^l_{h'} \times d^l_{w'}}$ is the sign of each element in $Q^{l+1}(\bar\x)$. Considering the impact of Rademacher variable $\sigma_i$ and similar to the method in FC layer, random variables $S' \in \{\pm 1\}^{ d^l_{h} \times d^l_{w}}$ and $U \in \R ^{d^l_c \times d^l_{h'} \times d^l_{w'}}$ are introduced to simply Eq.~(\ref{eq-conv-dev}), which are used to approximate $S$ and the channel selection process of $\K^{l+1}$ respectively.  Each element in $S'$ is $\pm1$ with equal probability and each element in $U$ follows the standard normal distribution $\mathcal N (0,1)$.  Given the gradient, the distortion $\bm \varepsilon^l_i$ is updated in a similar way as FC layer. The algorithm of the feature distortion on the convolutional layers is similar to Algorithm \ref{alg}. 

Different from the method applied on FC layers where each element of the binary mask $M^l$ is sampled independently, we draw lessons from DropBlock \cite{ghiasi2018dropblock} where elements in a contiguous square block with given size $block\_size$ of the feature map is distorted simultaneously. We denote the extension of the proposed method to convolutional layers as ``block feature map distortion''. 
\section{Experiments}
In this section, we conduct experiments on several benchmark datasets to validate the effectiveness of the proposed feature map distortion method. The method is implemented on both FC layers and convolutional layers, which are validated with conventional CNNs and modern CNNs (\eg ResNet) respectively. In order to set unified hyper-parameters $\gamma$ for different layers, we multiply $\gamma$ by the standard deviation of the feature maps in each layer, and alternately  update the distortion and weight one step for efficiency. The  distortion probability  (dropping probability for dropout and dropblock) increases linearly from 0 to the appointed distortion probability $p$ following \cite{ghiasi2018dropblock}. 
\subsection{Experiments on Fully Connected Layers}
\begin{table*}[t]
	\caption{Accuracies of ResNet-56 on CIFAR10 and CIFAR-100 dataset.}
		\vskip 0.1in
	\label{tb-res56}
	\centering
	\begin{tabular}{l||c|c}
		\hline
		Model & CIFAR-10 (\%) & CIFAR-100 (\%)\\ \hline\hline
		Resnet-56 &93.95 $\pm$ 0.09 & 71.81 $\pm$ 0.21\\
		Resnet-56 + DropBlock \cite{ghiasi2018dropblock}&94.18 $\pm$ 0.14 & 73.08  $\pm$ 0.23 \\
		\hline
		Resnet-56 + Block Feature Map Distortion &\textbf{94.50 $\pm$ 0.15}&\textbf{73.71 $\pm$ 0.20}\\ 
		\hline
	\end{tabular}	
\end{table*}
To validate the effect of the proposed feature map distortion method implemented on the FC layers, we conduct experiments on a conventional CNN on CIFAR-10 and CIFAR-100 dataset. The proposed method is compared with multiple state-of-the-art variants of dropout. 

\textbf{Dataset. } 
CIFAR-10 and CIFAR-100 dataset both contain 60000 natural images with size $32\times 32$. 50000 images are used for training and 10000 for testing. The images are divided into 10 categories and 100 categories, respectively. 20\% of the training data are regarded as validation sets. Data augmentation method is not used for fair comparison. 

\textbf{Implementation details. }
The conventional CNN has three convolutional layers with 96, 128 and 256 filters, respectively. Each layer consists of a $5\times 5$ convolutional operation with stride 1 followed by a $3\times3$ max-pooling operation with stride 2. Then the features are sent to two fully-connected layers with 2048 hidden units each. We implement the distortion method on each FC layer.  Distortion probability $p$ is selected from \{0,4, 0.5, 0.6\} and the step length $\gamma$ is set to 5. The model is trained for 500 epoch with batchsize 128. The learning rate is initialized with 0.01, and decayed by a factor of 10 at 200, 300 and 400 epochs. We run our method 5 times with different random seeds and report the average accuracy with standard deviation.

\textbf{Compared methods.} 
The  CNN model trained without extra regularization tricks is used as the baseline model. Further more, we compare our method with the widely used dropout method \cite{hinton2012improving} and several state-of-the-art variants, including Vardrop \cite{kingma2015variational}, Sparse Vardrop \cite{molchanov2017variational} and RDdrop \cite{zhai2018adaptive}.

\textbf{Results.}
The test accuracies on both CIFAR-10 and CIFAR-100 are summarized in Table \ref{tb-ccnnres}. The proposed feature map distortion method is superior to the compared methods by a large margin on both two datasets. CNN trained with the help of the proposed method achieves an accuracy of 85.24\%, which improves the performance of the state-of-the-art RDdrop method with 2.13\% and 1.58\% on CIFAR-10 and CIFAR-100 dataset, respectively. It shows that the proposed feature map distortion method can reduce the empirical Rademacher complexity effectively while preserve the representation power of the model, resulting in a better test performance.

\subsection{Experiments on Convolutional Layers}
It is much important to apply the proposed method to convolutional layer since modern CNN such as ResNet mostly consist of convolutional layers. In this section, we apply the proposed method on convolutional layers and conduct several experiments on both CIFAR-10 and CIFAR-100 dataset.

\textbf{Implementation details.} 
The widely-used ResNet-56 \cite{he2016identity} which contains three groups of blocks is used as the baseline model. DropBlock method \cite{ghiasi2018dropblock} is used as the peer competitor. Both the proposed block feature map distortion method and DropBlock method are implemented after each convolution layers in the last group with \textit{block\_size=6}, and the distortion probability (dropping probability for DropBlock) $p$ is selected from $\{0.01, 0.02, \cdots, 0.1\}$. The step length $\gamma$ is set to 30 empirically. Standard data augmentation including random cropping, horizontal flipping and rotation(within $\pm$15 degrees) are conducted during training. The networks are trained for 200 epochs, batchsize is set to 128 and weight decay is set to 5e-4. The initial learning rate is set to 0.1 and is decayed by a factor of 5 at 60, 120 and 160 epochs. All the methods are repeated 5 times with different random seeds and the average accuracies with standard deviations are reported.

\textbf{Results.}
The results on both CIFAR-10 and CIFAR-100 dataset are shown in Table \ref{tb-res56}. The proposed method is superior to DropBlock method and improves the performance with 0.32\% and 0.63\%, respectively. It shows that the proposed feature map distortion methods suits for convolutional layers and can improves the performance of modern network structures.

\begin{figure}[t]
	\centering
	\includegraphics[width=0.8\columnwidth]{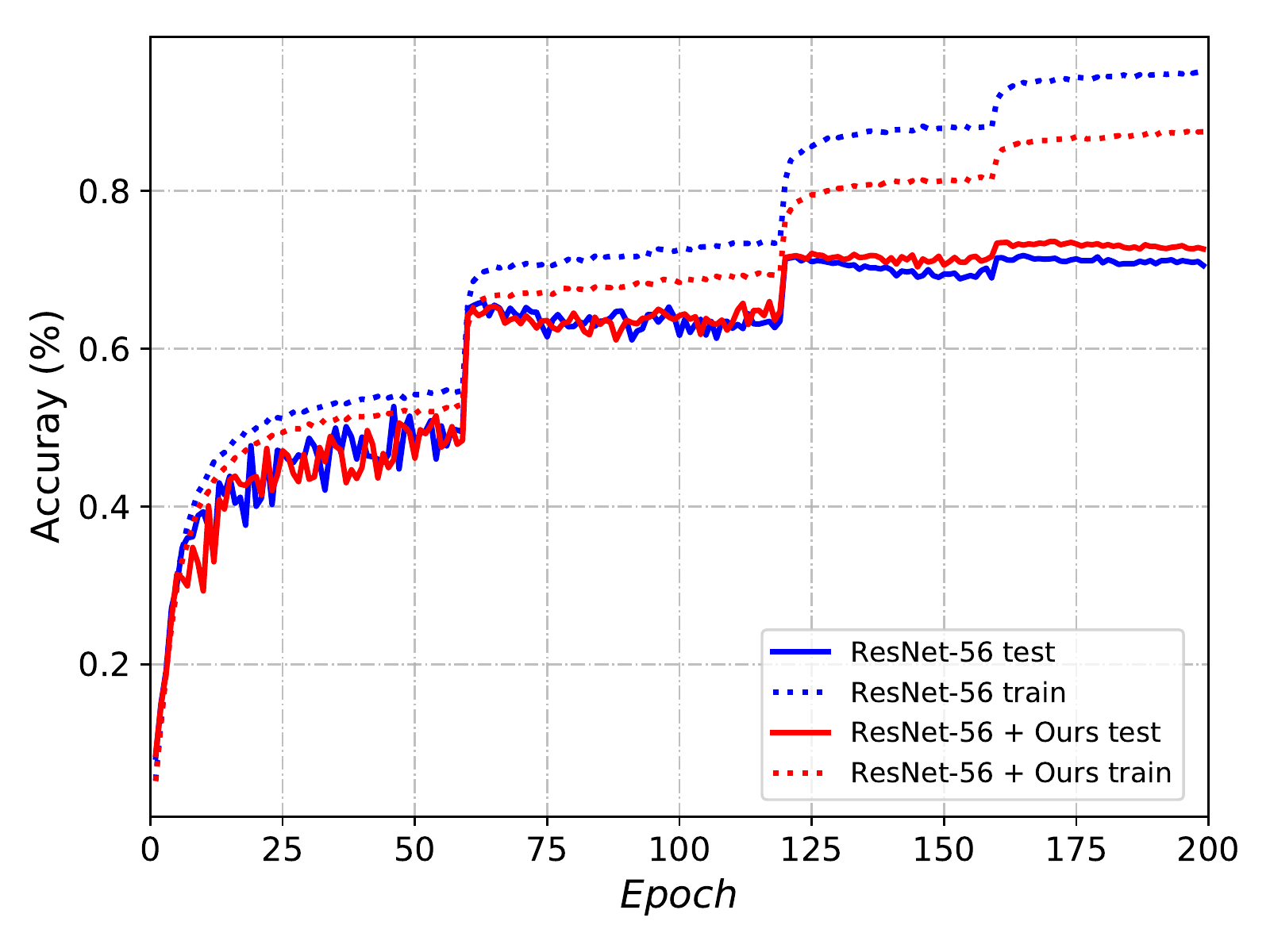}
	\vspace{-1em}
	\caption{Training curves on the CIFAR-100 dataset.}
	\label{fg-cur}
	\vspace{-1em}
\end{figure}

\textbf{Training curve.} The training curves on CIFAR-100 dataset are shown in Figure \ref{fg-cur}. The solid line and dotted line denote the test stage and the training stage respectively, while the red line and blue line denote the proposed feature map distortion method and the baseline model. When training converges, the baseline ResNet-56 traps in over-fitting problem and achieves a higher training accuracy but lower test accuracy, while the proposed feature map distortion method overcome this problem and achieves a higher test accuracy, which shows the improvement of model generalization ability.

\begin{table*}[t]
	\caption{Accuracies of ResNet-50 on ImageNet dataset.}
		\vskip 0.1in
	\label{tb-imgres}
	\centering
	\begin{tabular}{l||c|c}
		\hline
		Model & Top-1 Accuracy (\%) & Top-5 Accuracy (\%) \\ \hline
		ResNet-50 &76.51 $\pm$ 0.07&93.20 $\pm$ 0.05\\
		ResNet-50 + Dropout \cite{srivastava2014dropout} &76.80 $\pm$ 0.04&93.41 $\pm$ 0.04\\
		ResNet-50 + DropPath \cite{larsson2016fractalnet} &77.10 $\pm$ 0.08&93.50 $\pm$ 0.05\ \\
		ResNet-50 + SpatialDropout \cite{tompson2015efficient}&77.41 $\pm$ 0.04&93.74 $\pm$ 0.02\\
		ResNet-50 + Cutout \cite{devries2017improved}&76.52 $\pm$ 0.07&93.21 $\pm$ 0.04\\ 
		ResNet-50 + AutoAugment \cite{cubuk2018autoaugment}&77.63&93.82\\
		ResNet-50 + Label Smoothing \cite{szegedy2016rethinking}&77.17 $\pm$ 0.05&93.45 $\pm$ 0.03\\
		ResNet-50 + DropBlock \cite{ghiasi2018dropblock} &78.13 $\pm$ 0.05&94.02 $\pm$ 0.02\\ \hline
		ResNet-50 + Feature Map Distortion&{77.71 $\pm$ 0.05}&{93.89 $\pm$ 0.04}\\
		ResNet-50 + Block Feature Map Distortion&\textbf{78.76 $\pm$ 0.05}&\textbf{94.33 $\pm$ 0.03}\\
		\hline
	\end{tabular}	
\end{table*}

\begin{figure*}[htb] 
	\subfigure[]{
		\begin{minipage}[t]{0.33\linewidth}
			\centering
			\includegraphics[width=0.99\linewidth]{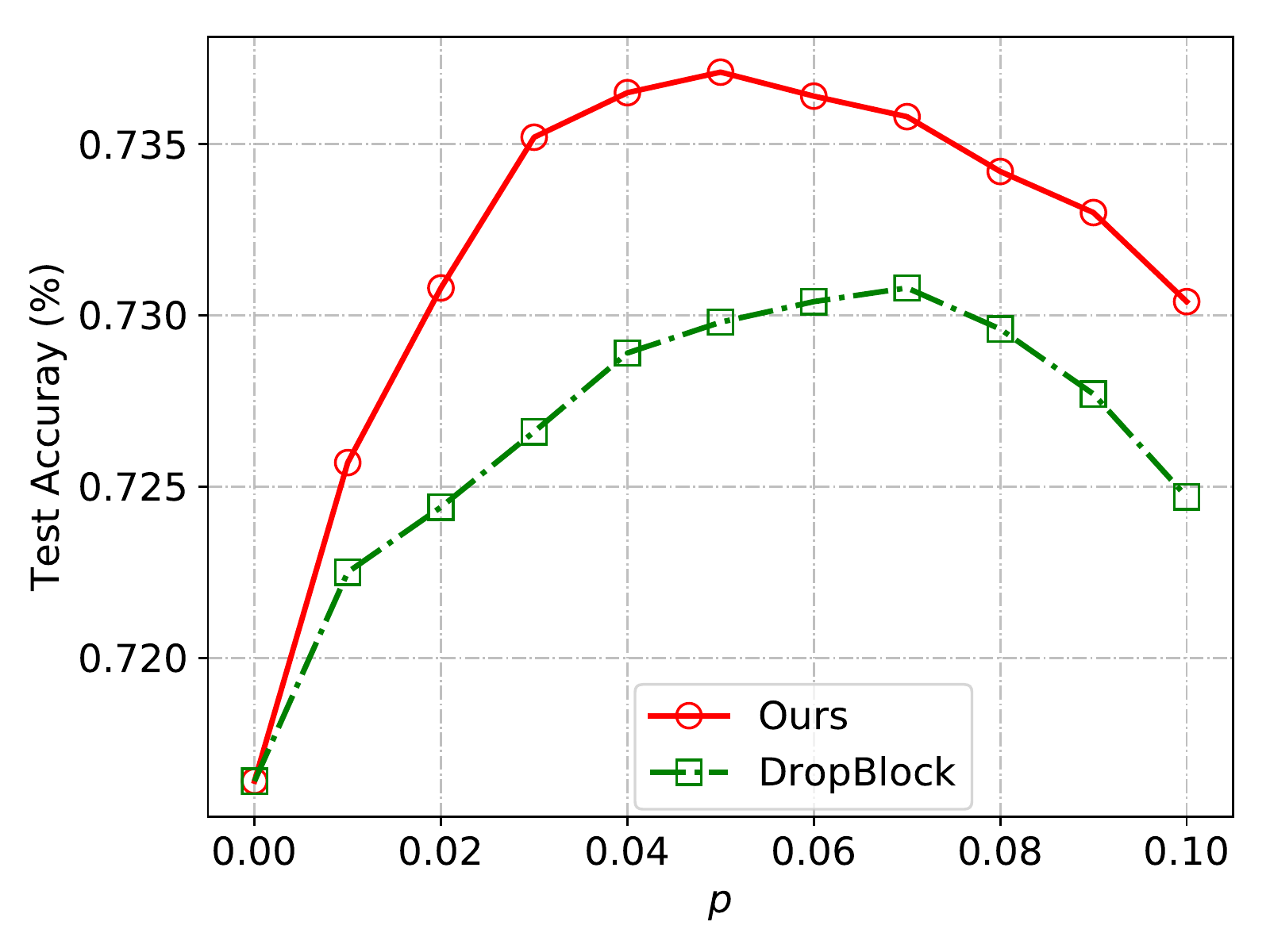}
		\end{minipage}
	}
	\subfigure[]{
		\begin{minipage}[t]{0.33\linewidth}
			\centering
			\includegraphics[width=0.99\linewidth]{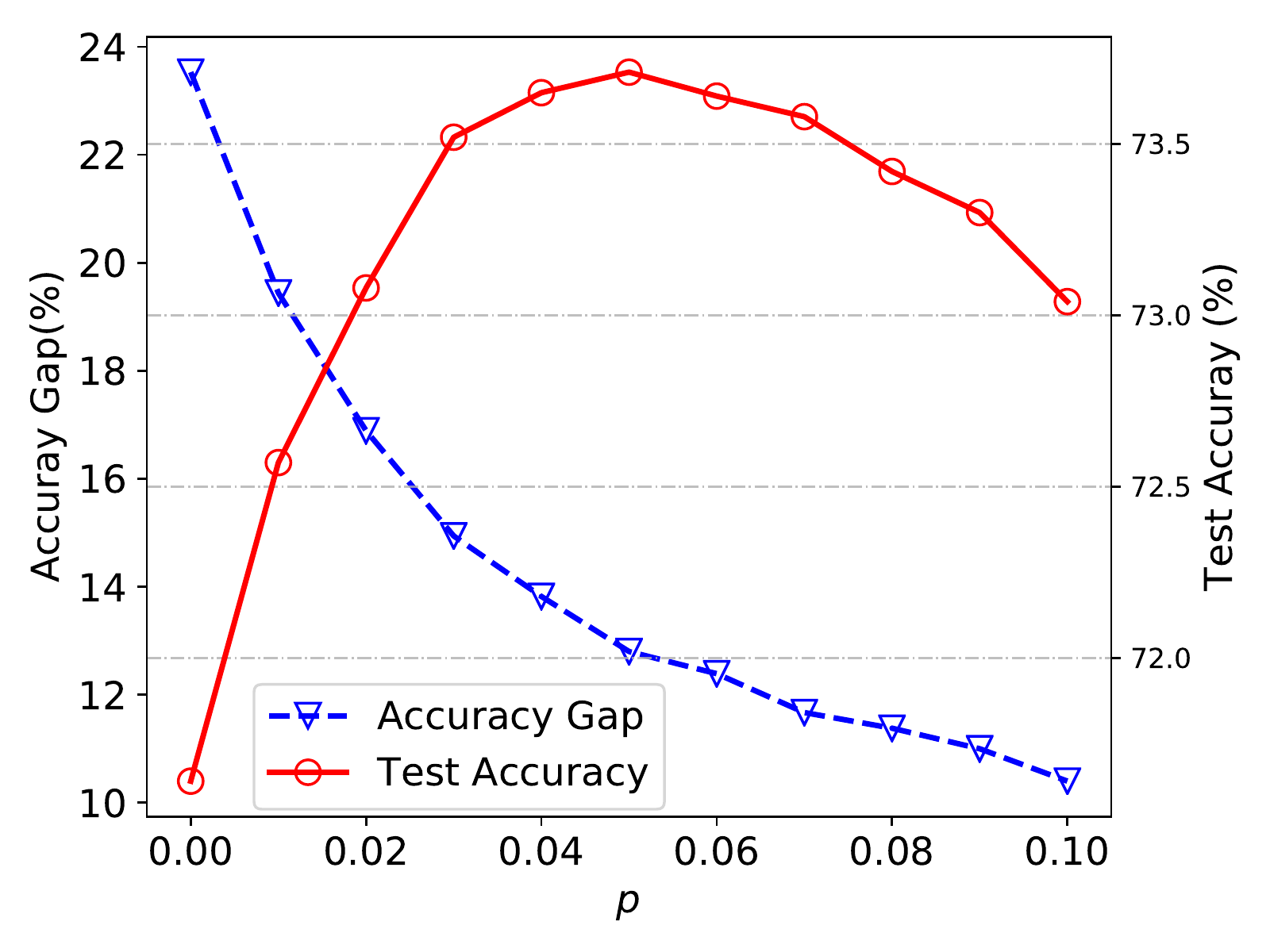}
		\end{minipage}
	}
	\subfigure[]{
		\begin{minipage}[t]{0.33\linewidth}
			\centering
			\includegraphics[width=0.99\linewidth]{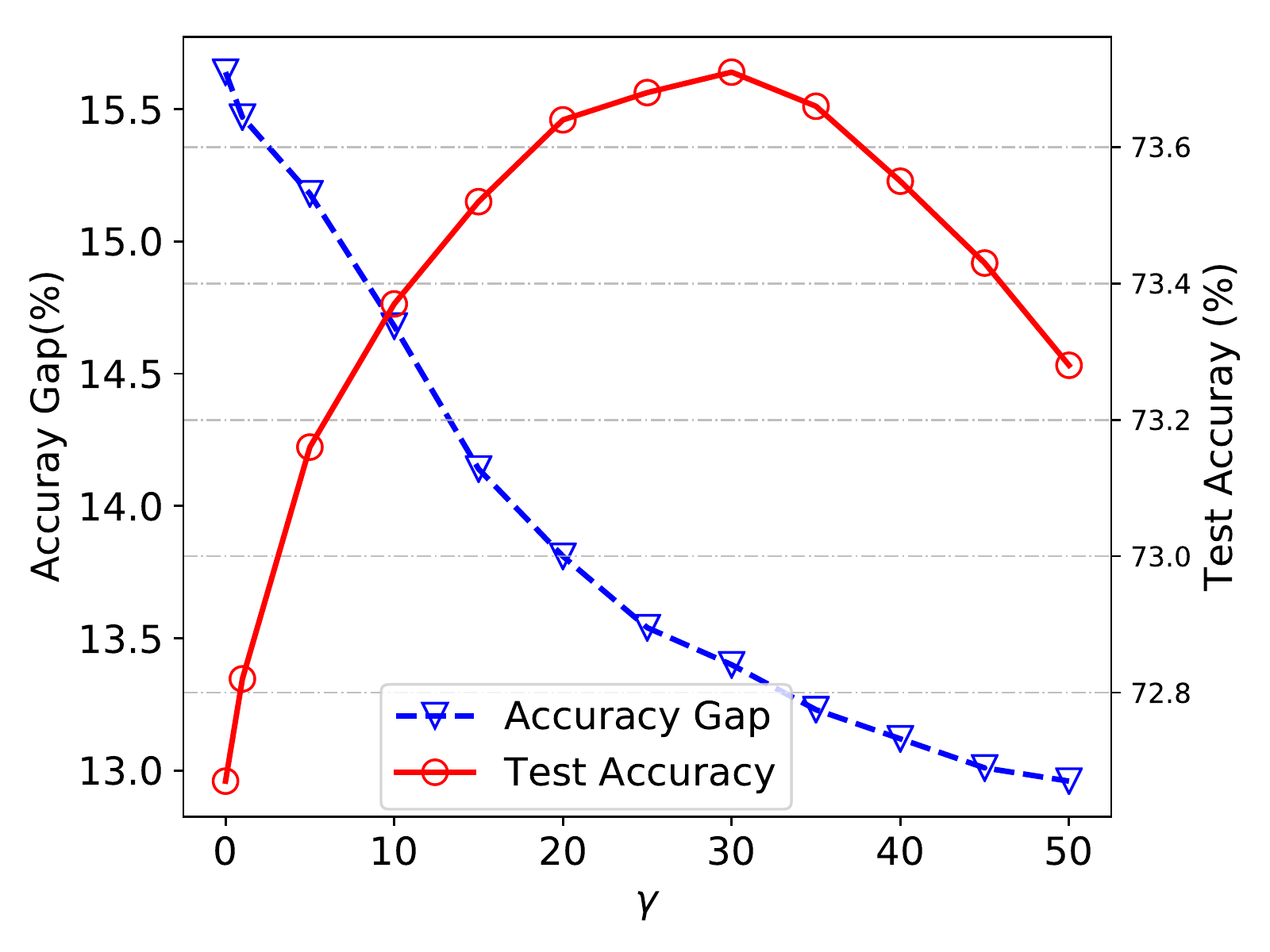}
		\end{minipage}
	}	
		\vspace{-2mm}
	\caption{The impact of distortion probability $p$ and step length $\gamma$ on CIFAR-100 dataset. Test accuracies \wrt distortion probability $p$ for feature map distortion and dropblock are shown in (a). Test accuracies and accuracy gaps \wrt distortion probability $p$ and step length $\gamma$ are shown in (b) and (c).}
	\label{fg-pbgamma}
	
		\vskip -0.1in
\end{figure*}

\textbf{Feature map distortion \emph{v.s.} DropBlock.} The test accuracy of our method (red) and the Dropblock method (green) with various distortion probability (dropping probability) $p$ on CIFAR-100 dataset are shown in Figure \ref{fg-pbgamma}(a). Increasing the drop probability $p$ enhances the effect of regularization, and the test accuracy can be improved when setting $p$ in an appropriate range. Note that our method achieves a better performance than DropBlock with $p$ in a larger range, which demonstrate the superior of feature map distortion. 

\textbf{Test accuracy \emph{v.s.} accuracy gap.} Figure \ref{fg-pbgamma}(b) and (c) show how test accuracy (red) and the accuracy gap between training and testing accuracies (blue) vary when setting different distortion probability $p$ and length step $\gamma$.  Larger $p$ implies that more locations of the feature maps are distorted while $\gamma$ controls the intensity of disturbing in each location. Increasing either $p$ or $\gamma$ bring stronger regularization, resulting in smaller gap between the training and testing accuracies, which means a stronger generalization ability. However, disturbing too many locations or disturbing a location with too much intensity may destroy the representation power and having negative impact on the final testing accuracy.
 Instead of using fixed intensity in conventional dropout and DropBlock method, out method applies proper intensity distortion on proper locations and results in better performance.

\subsection{Experiments on ImageNet Dataset}

In this section, we conducts experiments on large-scale ImageNet dataset and implement the feature map distortion method with conventional dropout and the recent DropBlock method, namely ``Feature Map Distortion'' and ``Block Feature Map Distortion'', respective.  

\textbf{Dataset.}
ImageNet dataset contains 1.2M training images and 50000 validation images, consisting of 1000 categories. Standard data augmentation methods including random cropping and horizontally flipping is conducted on training data. 

\textbf{Implementation details.} 
We follow the experimental settings in \cite{ghiasi2018dropblock} for fair comparison. The prevalent ResNet-50 is used as the baseline model. The distortions are applied on the feature maps after both convolutional layers and skip connections in the last two groups. The step length is set to $5$. For feature map distortion implemented based on conventional dropout, distortion probability $p$ (dropping probability) is set to 0.5 as suggested by \cite{srivastava2014dropout}. For Block feature map distortion, the \textit{block\_size} and $p$ (dropping probability) are set to 6 and 0.05 following \cite{ghiasi2018dropblock}. We report the single-crop top-1 and top-5 accuracies on the validation set and repeat the methods three time with different random seeds. 

\textbf{Compared method.} Multiple state-of-the-art regularization methods are compared, including dropout based methods, data augmentation and label smoothing. DropPath\cite{larsson2016fractalnet}, SpatialDropout\cite{tompson2015efficient} and  Dropblock \cite{ghiasi2018dropblock} are the state-of-the-art variants of dropout. Data augmentation including Cutout \cite{devries2017improved}  and AutoAugment \cite{cubuk2018autoaugment}), and label smoothing \cite{szegedy2016rethinking} are prevalent regularization techniques to alleviate over-fitting.

\textbf{Results.} 
In Table~\ref{tb-imgres}, the proposed feature distortion method can not only increase the performance of deep neural networks using conventional dropout method, but also enhance the peformance of the recent Dropblock method, since our method is also suitable and well adapted to convolutional layers. As a result, the feature map distortion improve the accuracy from 76.80\% to 77.71\% compared to the conventional dropout method . The block feature map distortion method achieves top-1 accuracy 78.76\%, which surpass other state-of-the art methods from a large margin. The results demonstrate that our method can simultaneously increase the generalization ability and preserving the useful information of original features.

\vspace{-2mm}

\section{Conclusion}
Dropout based methods have been successfully used for enhancing the generalization ability of deep neural networks. However, eliminating some of units in neural networks can be seen as a heuristic approach for minimizing the gap between expected and empirical risks of the resulting network, which is not the optimal one in practice. Here we propose to embed distortions onto feature maps of the given deep neural network by exploiting the Rademacher complexity. We further extend the proposed method to convolutional layers and explore the detailed feed-forward and back-propagation procedures. Thus, we can employ the proposed method into any off-the-shelf deep neural architectures. Extensive experimental results show that the feature distortion technique can be easily embedded into mainstream deep networks to achieve better performance on benchmark datasets over conventional approaches.

\section{ Acknowledgments}
This work is supported by National Natural Science Foundation of China under Grant No. 61876007, 61872012 and Australian Research Council under Project DE-180101438.
{\small
\bibliographystyle{aaai}
\bibliography{reference}}
\end{document}